\documentclass{article}
\usepackage[main, final]{neurips_2026}

\makeatletter
\renewcommand{\@notice}{}
\makeatother

\usepackage[utf8]{inputenc}
\usepackage[T1]{fontenc}
\usepackage{amsmath,amssymb,amsthm}
\usepackage{mathtools}
\usepackage{bm}
\usepackage{graphicx}
\usepackage{booktabs}
\usepackage[hidelinks]{hyperref}
\usepackage{url}
\usepackage{xcolor}

\newcommand{\R}{\mathbb{R}}
\newcommand{\E}{\mathbb{E}}

\newtheorem{lemma}{Lemma}[section]
\newtheorem{theorem}{Theorem}[section]
\newtheorem{corollary}{Corollary}[section]
\newtheorem{proposition}{Proposition}[section]
\newtheorem{remark}{Remark}[section]

\title{Robust Diffusion Models via Divergence-Induced Weighted Denoising}

\author{%
  Lei Li\thanks{Corresponding author: \texttt{lei.li@lunarai.llc}} \\
  LunarAI LLC \\
  Newark, DE 19702 \\
  \And
  Yuexiao Dong \\
  Fox School of Business \\
  Temple University \\
}

\begin{document}
\maketitle

\begin{abstract}
We show that replacing the standard MSE denoising loss in diffusion models with a nonlinear transformation induced by an f-divergence yields a simple robust training surrogate that empirically improves performance under data contamination, with small additional computational overhead.
The theoretical foundation rests on a local divergence construction: under the Gaussian reverse-kernel structure of DDPM, each per-step likelihood ratio follows a lognormal distribution parameterized by a scalar mismatch, so the conditional f-divergence at each step reduces to a one-dimensional function of the denoising error.
Summing these local divergences yields a training objective that unifies diffusion training as divergence-induced weighted denoising, where the derivative of the induced divergence acts as a residual-space influence weight that controls the contribution of each sample.
Bounded-influence divergences (Hellinger, negative exponential) suppress large-error samples, with Hellinger yielding an explicit exponential weight, connecting the framework to robust M-estimation.
Empirically, on CIFAR-10 under 30\% contamination, NED reduces FID from 93.0 (KL) to 77.5, while also outperforming standard robust losses such as Huber and clipped MSE.
\end{abstract}

\section{Introduction and Related Work}

Diffusion models have become a dominant paradigm for generative modeling, achieving state-of-the-art performance across a wide range of tasks including image, audio, and multimodal generation. Despite their empirical success, the standard training objective, a weighted mean squared error on the denoising residual, treats every training sample equally regardless of how corrupted or atypical it may be. When training data contains corrupted or low-quality examples, this equal weighting degrades generative quality.

In this paper we show that a simple, principled modification to diffusion training can provide robustness to such contamination. The key idea is to replace the standard per-step MSE loss $m_t = w_t\|\epsilon - \epsilon_\theta\|^2$ with a nonlinear transformation $h_G(m_t)$, where the function $h_G$ is induced by an f-divergence. In practice, this amounts to reweighting each sample's gradient by an influence weight $h_G'(m_t)$ that downweights large residuals. For Hellinger divergence, the reweighting is simply $e^{-m_t/4}$; for the negative exponential divergence (NED), it further suppresses high-mismatch samples. The modification adds small computational overhead and requires changing only one line of code. On CIFAR-10 with 30\% data contamination, this single change reduces FID from $93.0$ (standard KL/MSE) to $77.5$ (NED), while also yielding lower cross-seed variance.

These practical gains rest on a statistical framework that we develop in this paper. We begin with a path-space formulation of the forward and reverse processes. By lifting the marginal divergence between data and model distributions to the joint path space, we obtain a tangent upper bound via the data-processing inequality. Under the Gaussian reverse-kernel structure commonly used in diffusion models, we show that the per-step likelihood ratio follows a lognormal distribution parameterized by a scalar mismatch $m_t$. This motivates a local conditional training objective that takes the form of a sum of scalar divergences:
\[
\sum_{t} \E\!\left[h_G(m_t)\right],
\]
where $h_G$ is a one-dimensional function determined entirely by the choice of f-divergence $G$. The standard MSE objective corresponds to KL divergence ($h_{\mathrm{KL}}(m)=m$), recovering the DDPM mean-matching term (up to endpoint contributions in the full ELBO), while alternative divergences induce different transformations of the same residual. The derivative $h_G'(m)$ acts as an influence weight in the sense of robust M-estimation, governing how strongly different mismatch levels contribute to optimization. This connects diffusion training to classical robustness theory and provides a mechanism for the contamination robustness of bounded-influence divergences.

Our work is related to several lines of research. Diffusion models were introduced in \cite{ho2020ddpm} and further generalized through score-based formulations \cite{song2021sde}; subsequent developments include improved architectures \citep{karras2022edm,nichol2021improved}, accelerated sampling \citep{song2020ddim,lu2022dpm}, and one-step generation \citep{salimans2022progressive,song2023consistency}. \cite{song2021maximum} connected score matching to a weighted combination of KL divergences but did not frame diffusion training as general divergence minimization at the path level. In the f-divergence literature, \cite{nowozin2016fgan} introduced critic-based minimization via the Fenchel dual; related density-ratio approaches include \cite{nguyen2010divergence,sugiyama2012density}. \cite{li2025divergence} developed a divergence-minimization framework for latent-structure models with monotone decrease and bounded influence functions. Our work specializes this to diffusion, where the Gaussian structure yields an explicit lognormal likelihood ratio without a learned critic. In flow matching, \cite{lipman2023flowmatching} learns a velocity field via quadratic loss; our framework suggests extending such objectives to $h_G(\|v_\theta - v^*\|^2)$ (empirical validation left to future work). Concurrent work by \cite{xu2025fdistill} applies f-divergences to one-step distillation; our framework provides the path-space foundation and extends to general diffusion training.

Our main contributions are as follows. First, we establish a lognormal characterization of the per-step likelihood ratio under Gaussian reverse kernels (Lemma~\ref{lem:lognormal}), which, to our knowledge, has not been explicitly stated in the diffusion literature. Second, we construct a local conditional f-divergence objective for diffusion training, motivated by a path-space upper bound via the data-processing inequality (Theorem~\ref{thm:marginal_to_path}). Under explicit moment, orthogonality, and tail conditions, we provide an asymptotic product-to-sum reduction that justifies the resulting sum of scalar divergences (Theorem~\ref{thm:path_to_local_clean}, Corollary~\ref{cor:diffusion_clean}). Third, we characterize the local expansion, monotonicity, and robustness properties of the induced divergence (Theorems~\ref{thm:local_convergence},~\ref{thm:global_monotonicity}), including the decay behavior of the NED influence weight, yielding a unified interpretation of diffusion training objectives as divergence-induced weighted denoising, with a suggested extension to flow matching. Finally, we validate the framework on CIFAR-10 and CIFAR-100, showing that HD and NED outperform KL and standard robust losses under data contamination.

The remainder of the paper is organized as follows. Section~\ref{sec:upper_bounds} develops the divergence framework, including the lognormal characterization, path-space bound, local conditional construction, and robustness analysis. Section~\ref{sec:experiments} presents experimental results on CIFAR-10 and CIFAR-100. Section~4 discusses limitations and future directions. Proofs and additional experiments are in the appendix.

\section{A Divergence Framework for Diffusion Models}
\label{sec:upper_bounds}

\subsection{Background: diffusion models}

Our framework builds on the standard diffusion model setup, which we summarize here to establish notation. A diffusion model defines a forward process that progressively adds Gaussian noise to data $x_0\sim g$ over $T$ steps,
\[
q(x_t\mid x_{t-1}) = \mathcal N(x_t;\, \sqrt{1-\beta_t}\, x_{t-1},\, \beta_t I),
\]
with a noise schedule $\{\beta_t\}_{t=1}^T$. For large $T$, the terminal distribution $q(x_T)$ is approximately $\mathcal N(0,I)$. Generation proceeds by learning a reverse process $p_\theta(x_{t-1}\mid x_t) = \mathcal N(x_{t-1};\, \mu_{\theta,t}(x_t),\, \Sigma_t)$, where $\Sigma_t$ is typically fixed to $\beta_t I$ or $\tilde\beta_t I$ \citep{ho2020ddpm}. In the continuous-time limit, the forward process becomes an It\^o SDE and the reverse process is characterized by a score function $\nabla_{x_t}\log q(x_t)$, which is approximated by a neural network trained via denoising score matching \citep{song2021sde}. Under the standard $\epsilon$-parameterization, the network $\epsilon_\theta(x_t,t)$ predicts the noise component, and the training objective reduces to a weighted mean squared error $\sum_t w_t \|\epsilon - \epsilon_\theta(x_t,t)\|^2$.

Our framework is built on the general class of $f$-divergences. Given two distributions $P$ and $Q$ with $P\ll Q$, the $f$-divergence generated by a convex function $G:[-1,\infty)\to\mathbb R$ with $G(0)=0$ is defined as $D_G(P\|Q) = \mathbb E_Q[G(dP/dQ - 1)]$. When $G(\delta)=(1+\delta)\log(1+\delta)-\delta$ this recovers the KL divergence; $G(\delta)=2(\sqrt{1+\delta}-1)^2$ gives the squared Hellinger distance; and $G(\delta)=e^{-\delta}-1+\delta$ defines the negative exponential divergence (NED) \citep{basu1997minimum}. The normalization $G'(0)=0$, $G''(0)=1$ is adopted throughout.

A key structural observation that underlies our framework is the following characterization of the per-step likelihood ratio.

\begin{lemma}[Lognormal likelihood ratio]\label{lem:lognormal}
Consider the DDPM variational posterior $q_t(x_{t-1}\mid x_t,x_0)=\mathcal N(\tilde\mu_t,\Sigma_t)$ and a model reverse kernel $p_{\theta,t}(x_{t-1}\mid x_t)=\mathcal N(\mu_{\theta,t},\Sigma_t)$ with matched covariance. Define the per-step mismatch $m_t:=\tfrac12(\tilde\mu_t-\mu_{\theta,t})^\top\Sigma_t^{-1}(\tilde\mu_t-\mu_{\theta,t})$. Then, conditional on $(x_0,x_t)$, the per-step likelihood ratio $r_t:=q_t/p_{\theta,t}$ evaluated under the local model kernel $p_{\theta,t}(\cdot\mid x_t)$ satisfies
\[
\log r_t \sim N(-m_t,\, 2m_t).
\]
In particular, $r_t$ is lognormal with $\E[r_t\mid x_0,x_t]=1$ and $\mathrm{Var}(r_t\mid x_0,x_t)=e^{2m_t}-1$.
\end{lemma}

The proof is a direct calculation (Appendix~\ref{app:proof_lognormal}). Although the lognormal form of the Gaussian likelihood ratio is classical, this characterization parameterized by $m_t$ has not, to our knowledge, been stated in the diffusion literature. Existing work computes per-step KL divergences without isolating the distributional form of $r_t$ \citep{ho2020ddpm,song2021maximum}. Lemma~\ref{lem:lognormal} is the structural bridge that enables our framework: once the path-space divergence is reduced to per-step terms (Section~\ref{sec:path_to_local}), the lognormal form makes each term a one-dimensional function of~$m_t$.

\subsection{Marginal divergence to full path divergence}

The natural goal of diffusion training is to minimize the divergence $D_G(g\|f_\theta)$ between the data distribution $g$ and the model marginal $f_\theta$. However, this marginal divergence is generally intractable because $f_\theta$ is defined only implicitly through the reverse chain. A natural approach is to lift the problem to the joint path space $x_{0:T}$, which we use as motivation for the local objective constructed below.

\begin{theorem}[Lifted path-space divergence is a tangent upper bound]
\label{thm:marginal_to_path}
Let $D_G$ be a general divergence, $g,f_\theta$ the marginal data/model distributions on $x_0$, and $Q,P_\theta$ the corresponding lifted path laws on $x_{0:T}$, with $Q\ll P_\theta$ (otherwise the divergence may be infinite). Then $D_G(g\|f_\theta) \le D_G(Q\|P_\theta)$ for every~$\theta$, with equality when $Q(x_{1:T}\mid x_0)=P_\theta(x_{1:T}\mid x_0)$ for $g$-a.e.\ $x_0$. In particular, if $\theta_0$ denotes an exact lifted model satisfying $P_{\theta_0}=Q$ (equivalently, the above conditional-path equality), then $D_G(g\|f_{\theta_0})=D_G(Q\|P_{\theta_0})$.
\end{theorem}

The proof is given in Appendix~\ref{app:proof_marginal_to_path}.

Theorem~\ref{thm:marginal_to_path} shows that the complete-data divergence \(D_G(Q\|P_\theta)\) is a tangent upper bound to the marginal divergence \(D_G(g\|f_\theta)\): it dominates the marginal divergence for all \(\theta\), and becomes exact when the model reproduces the full path law. For KL divergence, this yields the standard ELBO, where the path likelihood ratio factorizes exactly into per-step terms plus endpoint contributions. For general f-divergences, the path likelihood ratio contains an additional endpoint factor $g(x_0)q(x_T\mid x_0)/p(x_T)$ that does not separate cleanly from the per-step product under a nonlinear~$G$. Rather than attempting to bound this factor, we use the path-space perspective as motivation and directly construct the training objective from the local conditional divergences at each step.

\subsection{Local conditional divergence construction}
\label{sec:path_to_local}

Rather than optimizing the full path-space divergence $D_G(Q\|P_\theta)$ directly (which, for general f-divergences, involves an endpoint factor that does not decompose into per-step terms), we construct the training objective from the local conditional f-divergences at each step. Specifically, we define
\[
\mathcal L_G(\theta)
:=
\sum_{t=1}^T
\E_{q(x_0,x_t)}\!\left[
D_G\!\left( q(\cdot\mid x_t,x_0) \;\|\; p_{\theta,t}(\cdot\mid x_t) \right)
\right].
\]
By Lemma~\ref{lem:lognormal}, the conditional divergence at each step is a function of the scalar mismatch $m_t = m_t(x_0,x_t)$:
\[
D_G\!\left( q(\cdot\mid x_t,x_0) \;\|\; p_{\theta,t}(\cdot\mid x_t) \right)
=
h_G(m_t),
\]
where $h_G(m):=\E[G(U_m-1)]$ with $\log U_m \sim N(-m,2m)$. Hence
\begin{equation}\label{eq:local_objective}
\mathcal L_G(\theta)
=
\sum_{t=1}^T \E_{q(x_0,x_t)}\!\left[ h_G(m_t(x_0,x_t)) \right].
\end{equation}
This objective is mathematically well-defined for any f-divergence $G$ and does not require the path likelihood-ratio factorization. For KL divergence, $h_{\mathrm{KL}}(m)=m$ and the objective recovers the standard DDPM denoising loss. For general f-divergences, $\mathcal L_G$ serves as a tractable training surrogate motivated by the path-space upper bound (Theorem~\ref{thm:marginal_to_path}).

The following results, whose proofs are in the appendix, provide asymptotic justification for the local construction by showing that, under regularity conditions, the sum of per-step divergences approximates the divergence of a product likelihood ratio.

\begin{theorem}[Product-to-sum reduction]
\label{thm:path_to_local_clean}
Let $\delta_t\ge -1$ for all $t$, $R_T:=\prod_{t=1}^T (1+\delta_t)$, $S_T:=\sum_{t=1}^T \delta_t$, $B_T:=R_T-1-S_T$.
Assume $a_T>0$ is deterministic with $a_T\to 0$, and $G\in C^3$ near $0$ with $G(0)=G'(0)=0$, $G''(0)=1$.
Suppose:
\emph{(A1)}~$\E[S_T^2] = O(a_T)$;\;
\emph{(A2)}~$\E[S_T^2] - \sum_t \E[\delta_t^2] = o(a_T)$;\;
\emph{(A3)}~$\E|S_T|^3 = o(a_T)$;\;
\emph{(A4)}~$\E|S_TB_T| = o(a_T)$ and $\E[B_T^2] = o(a_T)$;\;
\emph{(A5)}~$\E|B_T|^3 = o(a_T)$;\;
\emph{(A6)}~$\sum_t \E|\delta_t|^3 = o(a_T)$.
Additionally, assume $\E\!\left[|G(S_T{+}B_T)|\,\mathbf 1\{|S_T{+}B_T|>\eta\}\right]=o(a_T)$ and $\sum_t \E\!\left[|G(\delta_t)|\,\mathbf 1\{|\delta_t|>\eta\}\right]=o(a_T)$ for any $\eta>0$ (tail integrability).
Then
\[
\E\!\left[G\!\left(\textstyle\prod_{t=1}^T(1+\delta_t)-1\right)\right]
=
\sum_{t=1}^T \E[G(\delta_t)] + o(a_T).
\]
\end{theorem}

\begin{corollary}[Diffusion Gaussian specialization]
\label{cor:diffusion_clean}
Under the setting of Lemma~\ref{lem:lognormal}, let $\E_{\mathrm{loc}}$ denote expectation under the auxiliary joint construction in which, conditional on $(x_0,x_t)$, each likelihood-ratio increment $r_t$ is evaluated under $p_{\theta,t}(\cdot\mid x_t)$ as in Lemma~\ref{lem:lognormal}. Define $\delta_t:=r_t-1$ and $a_T:=\sum_{t=1}^T \E[m_t]$.
Assume:
\emph{(D1)}~$a_T\to 0$, $\max_t m_t \le c_T\to 0$ almost surely, $\sum_t \E[m_t^2]=o(a_T)$, and $\sum_t \E[m_t^{3/2}]=o(a_T)$;\;
\emph{(D2)}~the increments $\{\delta_t\}$ are pairwise uncorrelated: $\E_{\mathrm{loc}}[\delta_s\delta_t]=0$ for $s\ne t$;\;
\emph{(D3)}~$\E_{\mathrm{loc}}|S_T|^3=o(a_T)$;\;
\emph{(D4)}~with $R_T,S_T,B_T$ as in Theorem~\ref{thm:path_to_local_clean}, $\E_{\mathrm{loc}}|S_TB_T|=o(a_T)$, $\E_{\mathrm{loc}}[B_T^2]=o(a_T)$, $\E_{\mathrm{loc}}|B_T|^3=o(a_T)$;
and the tail integrability conditions of Theorem~\ref{thm:path_to_local_clean}.
Then
$\E_{\mathrm{loc}}\!\left[G(R_T{-}1)\right] = \sum_{t=1}^T \E_{\mathrm{loc}}\!\left[ h_G(m_t) \right] + o(a_T).$
\end{corollary}

The proof is given in Appendix~\ref{app:proof_corollary}. For KL divergence, $h_{\mathrm{KL}}(m)=m$ and the local conditional objective recovers the standard DDPM mean-matching term $\sum_t \E[m_t]$; in the full ELBO, this appears together with the usual endpoint and reconstruction terms. For general f-divergences, no such chain rule holds, and the first-order reduction provides asymptotic justification for the local construction~\eqref{eq:local_objective}. Detailed interpretations of the abstract conditions (A1)--(A6) in Theorem~\ref{thm:path_to_local_clean} are given in Appendix~\ref{app:proof_path_to_local}.

Condition~(D1) is the natural local regime in which the per-step mismatches are uniformly small ($\max_t m_t\to 0$) and spread across many steps, the expected setting for a well-trained diffusion model. Condition~(D2) is a pairwise uncorrelatedness assumption on the likelihood-ratio increments, used to eliminate cross terms in $S_T^2$; it is an approximation whose quality depends on the model architecture and training stage, and is most natural when each $\delta_t$ is small. Even when~(D2) is violated, the local objective~\eqref{eq:local_objective} remains well-defined and can be used for training; (D2) is only needed for the asymptotic justification. Condition~(D3) controls the third moment of $S_T$, ensuring no single increment dominates, and~(D4) controls the higher-order product remainder $B_T=R_T-1-S_T$; both are supported by the lognormal moment bounds from~(D1) but are imposed explicitly.

\begin{remark}[Local objective vs.\ path-space bound]\label{rem:tightness}
Since $h_G(m)=m+o(m)$ (Theorem~\ref{thm:local_convergence}), all divergences collapse to nearly the same value when $m_t$ is small (at $m{=}0.01$ the three divergences agree to four digits); the choice of divergence matters primarily in the large-mismatch regime where robustness is needed. The local construction is used for \emph{objective design}, not as a convergence guarantee; Section~\ref{sec:experiments} validates effectiveness at finite mismatch.
\end{remark}

We now unpack the practical content of this reduction.

\subsection{Divergence-Induced Weighted Denoising Objectives}

By the local conditional construction~\eqref{eq:local_objective}, the training objective takes the form
\[
\mathcal L_G(\theta)
=
\sum_{t=1}^T \E_{q(x_0,x_t)}\!\left[ h_G(m_t) \right],
\]
where \(m_t=m_t(x_0,x_t)\) is the one-step Gaussian mismatch. Under the standard $\epsilon$-parameterization, this mismatch takes the form
\[
m_t = w_t \|\epsilon - \epsilon_\theta(x_t,t)\|^2,
\qquad
w_t = \tfrac{\beta_t^2}{2\alpha_t(1-\bar\alpha_t)\sigma_t^2},
\]
where $\alpha_t=1-\beta_t$, $\bar\alpha_t = \prod_{s=1}^t(1-\beta_s)$, and $\sigma_t^2$ is the reverse-kernel variance. Note that $w_t$ depends only on the noise schedule, not on the divergence. (The formula above is specific to the $\epsilon$-parameterization; other parameterizations such as $x_0$- or $v$-prediction yield different~$w_t$.) In stochastic training, the objective is estimated by sampling $x_0,t,\epsilon$ and evaluating $h_G(m_t)$.
Concretely, for KL this reduces to the standard weighted MSE: $\mathcal L_{\mathrm{KL}} = \sum_t w_t\|\epsilon - \epsilon_\theta\|^2$. For HD, $\mathcal L_{\mathrm{HD}} = \sum_t 4(1 - e^{-m_t/4})$. For NED, $h_{\mathrm{NED}}(m)=\mathbb E[e^{1-U_m}]-1$ (no closed form; evaluated via quadrature, see Section~\ref{sec:robustness}). Thus different divergences induce different transformations of the same denoising residual; the exact form is governed by $h_G$.

To understand the optimization behavior, differentiate \(\mathcal L_G(\theta)\) with respect to \(\theta\):
\[
\nabla_\theta \mathcal L_G(\theta)
=
\sum_{t=1}^T
h_G'(m_t)\, w_t\, \nabla_\theta \|\epsilon - \epsilon_\theta(x_t,t)\|^2.
\]
Thus, the effective per-sample weight is
\[
h_G'(m_t)\, w_t.
\]
The quantity \(h_G'(m)\) therefore plays the role of an adaptive residual-space influence weight, analogous to the $\psi$-function in robust M-estimation. Near the origin, all smooth divergences satisfy \(h_G'(0)=1\), so the resulting objective agrees locally with the standard weighted MSE loss. Away from the origin, however, different choices of \(G\) induce different downweighting behavior for large mismatches.

This perspective is useful for interpreting classical divergences. For KL divergence,
\[
h_{\mathrm{KL}}(m)=m,
\qquad
h'_{\mathrm{KL}}(m)=1,
\]
so every sample contributes with the standard diffusion weight \(w_t\). For HD,
\[
h_{\mathrm{HD}}(m)=4(1-e^{-m/4}),
\qquad
h'_{\mathrm{HD}}(m)=e^{-m/4},
\]
which yields exponential downweighting of large-error samples. For NED, the induced weight \(h'_{\mathrm{NED}}(m)\) also decays rapidly for large \(m\), leading to strong suppression of high-mismatch observations.

Therefore, diffusion training under a general divergence can be interpreted as a divergence-induced weighted denoising problem: the residual \(\|\epsilon-\epsilon_\theta(x_t,t)\|^2\) is unchanged, but the divergence determines how strongly different mismatch levels are emphasized.

\begin{table}[t]
\centering
\caption{Shifted generators, induced local divergences, and corresponding influence weights for the three classical divergences studied in this paper. Here \(U_m\) is the lognormal likelihood ratio of Lemma~\ref{lem:lognormal} and \(Z\sim N(0,1)\).}
\label{tab:generators_hg_weights}
\begin{tabular}{llll}
\toprule
Divergence & \(G(\delta)\) & \(h_G(m)\) & \(h_G'(m)\) \\
\midrule
KL
&
\((1+\delta)\log(1+\delta)-\delta\)
&
\(m\)
&
\(1\)
\\[4pt]
HD
&
\(2\big(\sqrt{1+\delta}-1\big)^2\)
&
\(4\big(1-e^{-m/4}\big)\)
&
\(e^{-m/4}\)
\\[4pt]
NED
&
\(e^{-\delta}-1+\delta\)
&
\(\mathbb E[e^{1-U_m}]-1\)
&
\(\mathbb E\!\left[e^{1-U_m}U_m\!\left(1{-}\tfrac{Z}{\sqrt{2m}}\right)\right]\)
\\
\bottomrule
\end{tabular}
\end{table}

Table~\ref{tab:generators_hg_weights} summarizes the shifted generators, the resulting local divergence maps, and their induced influence weights. KL has constant weight (no robustness); HD decays exponentially, while NED decays even more strongly asymptotically (Proposition~\ref{prop:ned_decay}); in practice, $h'_{\mathrm{NED}}$ is evaluated numerically by quadrature.

The resulting training algorithm (Figure~\ref{fig:algorithm}) differs from standard DDPM training by a single modification: the MSE loss $m_t$ is replaced by $h_G(m_t)$, or equivalently, each sample's gradient is reweighted by $h_G'(m_t)$. When $h_G$ and $h_G'$ admit closed forms (KL, HD), they are evaluated directly. When no closed form exists (e.g., NED), $h_G'(m)$ is approximated via Gauss--Legendre quadrature over the lognormal variable $U_m$ from Lemma~\ref{lem:lognormal}; our ablations (Appendix~\ref{app:experiments}) show that 16 quadrature points suffice.

\begin{figure}[t]
\centering
\small
\fbox{\parbox{0.92\textwidth}{%
\textbf{Divergence-Induced Diffusion Training}\\[4pt]
\textbf{Input:} dataset $\mathcal D$, divergence generator $G$, noise schedule $\{\beta_t\}$, learning rate $\eta$\\
\textbf{Output:} trained denoiser $\epsilon_\theta$\\[2pt]
\textbf{repeat}\\
\quad 1.\ Sample $x_0 \sim \mathcal D$,\; $t \sim \mathrm{Uniform}\{1,\dots,T\}$,\; $\epsilon \sim \mathcal N(0,I)$\\
\quad 2.\ Compute $x_t = \sqrt{\bar\alpha_t}\, x_0 + \sqrt{1-\bar\alpha_t}\,\epsilon$\\
\quad 3.\ Compute per-step mismatch $m_t = w_t \|\epsilon - \epsilon_\theta(x_t, t)\|^2$\\
\quad 4.\ Evaluate influence weight:\\
\qquad\quad \textbf{if} $h_G'$ has closed form (e.g., KL: $h_G'=1$;\; HD: $h_G'=e^{-m/4}$): evaluate directly\\
\qquad\quad \textbf{else} (e.g., NED): approximate $h_G'(m_t)$ via quadrature over $U_m$ (Lemma~\ref{lem:lognormal})\\
\quad 5.\ Update $\theta \leftarrow \theta - \eta\, h_G'(m_t)\, w_t\, \nabla_\theta \|\epsilon - \epsilon_\theta(x_t,t)\|^2$\\
\textbf{until} convergence
}}
\caption{Divergence-induced diffusion training. The only modification to standard DDPM is step~4: each sample's gradient is reweighted by the influence weight $h_G'(m_t)$.}
\label{fig:algorithm}
\end{figure}

\subsection{Properties of the Induced Divergence}

The local conditional construction shows that divergence-induced diffusion training can be expressed in terms of the scalar local divergence
\[
h_G(m)
=
\mathbb E\!\left[G(U_m-1)\right],
\qquad
\log U_m \sim N(-m,2m),
\]
where $U_m$ is the lognormal likelihood ratio from Lemma~\ref{lem:lognormal} and \(m\) is the per-step mismatch. We now characterize the local behavior of \(h_G(m)\) as \(m \to 0\).

\begin{theorem}[Local expansion of the induced divergence]
\label{thm:local_convergence}
Let $G\in C^3$ near $0$ with $G(0)=G'(0)=0$, $G''(0)=1$. Assume $\E[|G(U_m-1)|]<\infty$ for $m$ in a neighborhood of $0$, and the tail condition $\E[|G(U_m-1)|\,\mathbf{1}\{|U_m-1|>\eta\}]=O(m^{3/2})$ for some fixed $\eta>0$. Then $h_G(m) = m + O(m^{3/2})$ as $m\to 0$; in particular, $h_G(0)=0$ and $h_G'(0)=1$. For KL, HD, and NED, the sharper rate $h_G(m) = m + O(m^2)$ holds (with tail condition strengthened to $O(m^2)$), verified by direct expansion and a lognormal Chernoff bound (see Appendix~\ref{app:proof_local}).
\end{theorem}

The proof is given in Appendix~\ref{app:proof_local}.

\begin{theorem}[Global monotonicity of the induced divergence]
\label{thm:global_monotonicity}
Let $h_G(m)=\E[G(U_m-1)]$ with $U_m=\exp(-m+\sqrt{2m}\,Z)$, $Z\sim N(0,1)$, $m\ge 0$, where $G:[-1,\infty)\to\R$ is convex with $G(0)=G'(0)=0$. Assume $h_G(m)<\infty$ for all $m\ge 0$. Then $h_G$ is nonnegative and nondecreasing on $[0,\infty)$ with $h_G(0)=0$. If $G$ is strictly convex, then $h_G(m)>0$ for every $m>0$.
\end{theorem}

The proof is given in Appendix~\ref{app:proof_monotonicity}.

Theorem~\ref{thm:local_convergence} shows that all smooth divergences agree with KL to first order near $m{=}0$. The integrability condition $\E[|G(U_m{-}1)|]{<}\infty$ is verified for KL, HD, and NED by noting that $G(U_m{-}1)$ grows at most as $U_m\log U_m$ and all polynomial moments of the lognormal $U_m$ are finite (though $\E[e^{cU_m}]{=}\infty$ for $c{>}0$). Theorem~\ref{thm:global_monotonicity} complements this with a global result: $h_G$ increases monotonically with $m$. Together, divergences differ only in their higher-order behavior, which governs robustness. This is consistent with general divergence-minimization frameworks for latent-structure models \citep{li2025divergence}. Since $h_G'(0){=}1$, all divergence-induced objectives share the same first-order optimization geometry near convergence.

\subsection{Robustness Analysis of Classical Divergences}
\label{sec:robustness}

The induced objective $\mathcal L_G(\theta) = \sum_t \E[h_G(m_t)]$ makes clear that different divergences differ only through their influence weight $h_G'(m)$. We now examine what this means for KL, HD, and NED.

For KL divergence, \(h'_{\mathrm{KL}}(m)=1\), so all samples contribute equally regardless of mismatch magnitude. This corresponds to a non-robust objective: outliers and contaminated samples receive the same weight as clean data, making KL training sensitive to data corruption.

For HD,
\[
h'_{\mathrm{HD}}(m) = e^{-m/4},
\]
which decays exponentially with the mismatch \(m\). This provides robustness by strongly downweighting large-error samples, but such rapid decay may also suppress informative samples in early training when mismatches are large.

For the NED generator \(G_{\mathrm{NED}}(\delta)=e^{-\delta}-1+\delta\), the induced influence weight does not admit a closed form. Differentiating $h_{\mathrm{NED}}(m)=e\,\E[e^{-U_m}]-1$ under the integral sign gives
\[
h'_{\mathrm{NED}}(m)
=
e\,\mathbb E\!\left[
e^{-U_m}\,
U_m\!\left(1-\frac{Z}{\sqrt{2m}}\right)
\right],
\]
which decays to zero as $m\to\infty$. Although no closed form exists, we can establish the asymptotic rate.

\begin{proposition}[NED exponential decay]\label{prop:ned_decay}
As $m\to\infty$, $h'_{\mathrm{NED}}(m) \sim \frac{e}{4\sqrt{m}}\,e^{-m/4}$.
\end{proposition}

\begin{figure}[t]
\centering
\includegraphics[width=0.55\textwidth]{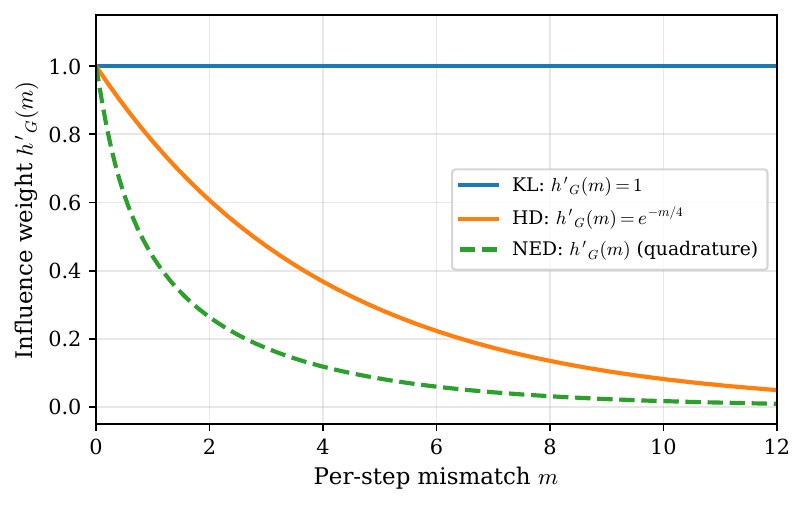}
\caption{Influence weights $h_G'(m)$~for KL, HD, and NED. KL assigns constant weight to all samples regardless of mismatch. HD and NED both decay exponentially at rate $e^{-m/4}$ (Proposition~\ref{prop:ned_decay}), with NED decaying more aggressively due to the polynomial correction.}
\label{fig:gradient_weights}
\end{figure}

The proof is given in Appendix~\ref{app:ned_decay}. Thus NED decays at the same exponential rate $e^{-m/4}$ as HD, with an additional $m^{-1/2}$ polynomial correction that makes NED slightly more aggressive. Figure~\ref{fig:gradient_weights} visualizes these influence weights.

In summary, KL provides no robustness, while HD and NED both achieve robustness through exponential suppression of large errors at rate $e^{-m/4}$. Whether this suppression is excessive depends on the contamination regime: for heavy contamination, aggressive downweighting is beneficial; for mild contamination or clean data, it may discard useful learning signal. This trade-off is governed entirely by the shape of $h_G'(m)$ and can be analyzed within the unified framework without heuristic loss design.

These observations suggest a broader design space of monotone, locally linear, saturating residual transformations $h$, even when they are not explicitly induced by a classical f-divergence. Exploring this design space is an interesting direction for future work.

\section{Experiments}
\label{sec:experiments}

We evaluate whether the theoretical robustness properties derived in Section~\ref{sec:upper_bounds} translate into measurable differences in generative quality under data contamination. All methods share identical architecture, hyperparameters, and training budget; only the divergence-induced loss function differs.

We use a 35M-parameter DDPM U-Net (base channels 128, channel multipliers $(1,2,2,2)$, self-attention at $16{\times}16$) with a linear $\beta_t$ schedule ($\beta_1{=}10^{-4}$, $\beta_T{=}0.02$), $T{=}1000$ forward steps, Adam optimizer at learning rate $2{\times}10^{-4}$, batch size $128$, and bfloat16 mixed-precision. All runs last 100K gradient steps. FID is computed with 5K generated samples against the clean test split via a 50-step DDIM sampler. The absolute FID values reflect this deliberately small model and short training budget, chosen for controlled comparison; the relative ordering across divergences is the quantity of interest. To simulate contamination, a fraction $\varepsilon$ of training images is replaced by a uniform mixture of four image-level corruption types ($8{\times}8$ patch noise, Gaussian blur $\sigma{=}1.5$, salt-and-pepper $p{=}0.05$, contrast shift $\pm 0.4$). Since the model is unconditional, label information is not used. Each setting is run with three seeds (42, 123, 456); we report mean $\pm$ standard deviation.

Table~\ref{tab:cifar_results} reports CIFAR-10 \citep{krizhevsky2009cifar} FID scores for the three divergence-induced objectives.

\begin{table}[t]
\centering
\caption{CIFAR-10 FID (lower is better) at 100K steps under $\varepsilon$-contamination, mean $\pm$ std over 3 seeds. Bold: best per row.}
\label{tab:cifar_results}
\begin{tabular}{lccc}
\toprule
$\varepsilon$ & KL & HD & NED \\
\midrule
$0.00$ & $35.4 \pm 12.3$ & $\bm{31.8 \pm 1.7}$ & $32.4 \pm 2.4$ \\
$0.10$ & $48.8 \pm 11.0$ & $40.5 \pm 5.7$ & $\bm{39.8 \pm 3.8}$ \\
$0.20$ & $67.7 \pm 32.6$ & $66.5 \pm 27.9$ & $\bm{62.2 \pm 18.3}$ \\
$0.30$ & $93.0 \pm 29.6$ & $91.9 \pm 33.4$ & $\bm{77.5 \pm 19.7}$ \\
\bottomrule
\end{tabular}
\end{table}

On clean data, all three divergences achieve comparable FID, consistent with the first-order equivalence $h_G(m)=m+O(m^2)$ (Theorem~\ref{thm:local_convergence}). Notably, HD and NED exhibit substantially lower cross-seed variance than KL ($1.7$ and $2.4$ vs.\ $12.3$), suggesting that bounded-influence objectives stabilize training. Under contamination ($\varepsilon{=}0.30$), NED achieves the best mean FID ($77.5$), improving over KL by $15.5$ points. The advantage of bounded-influence divergences grows with the contamination level, consistent with the theoretical prediction that the divergence choice matters primarily in the large-mismatch regime.

We also compare the divergence-induced objectives with two standard robust losses: Huber loss ($\delta{=}1.0$) and clipped MSE (threshold $\tau{=}10$). Table~\ref{tab:baselines} reports the results using seed~42 for a controlled single-seed comparison.

\begin{table}[t]
\centering
\caption{CIFAR-10 FID: divergence-induced objectives vs.\ standard robust losses (single seed, 100K steps). Bold: best per row.}
\label{tab:baselines}
\begin{tabular}{lcccc}
\toprule
$\varepsilon$ & Huber & ClipMSE & HD & NED \\
\midrule
$0.00$ & $\bm{27.8}$ & $\bm{27.7}$ & $32.4$ & $31.6$ \\
$0.10$ & $53.6$ & $54.0$ & $43.3$ & $\bm{42.3}$ \\
$0.20$ & $87.2$ & $67.5$ & $\bm{45.2}$ & $49.8$ \\
$0.30$ & $94.0$ & $90.5$ & $56.2$ & $\bm{55.1}$ \\
\bottomrule
\end{tabular}
\end{table}

At moderate to high contamination ($\varepsilon \ge 0.10$), both HD and NED substantially outperform Huber and clipped MSE. At $\varepsilon{=}0.30$, HD and NED achieve FID scores of $56.2$ and $55.1$, compared to $94.0$ and $90.5$ for the standard baselines, an improvement of $35{-}40\%$. This demonstrates that the divergence-induced weighting, which arises from the path-space framework, provides robustness that is not easily replicated by standard modifications of the MSE loss. We note that the Huber and clipping thresholds ($\delta{=}1.0$, $\tau{=}10$) were chosen as representative defaults; a grid search over these hyperparameters might narrow the gap. Systematic tuning of robust baselines is left for future work.

\begin{figure}[t]
\centering
\includegraphics[width=0.6\textwidth]{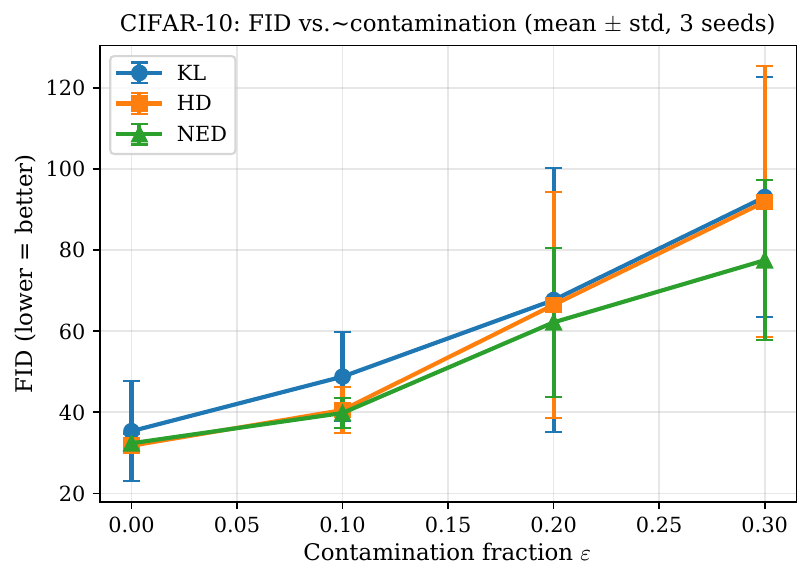}
\caption{CIFAR-10 FID vs.\ contamination level (mean $\pm$ std over 3 seeds). NED degrades least and maintains lower variance than KL, especially under contamination. All divergences are comparable on clean data, consistent with the first-order equivalence (Theorem~\ref{thm:local_convergence}).}
\label{fig:fid_vs_eps}
\end{figure}

Figure~\ref{fig:fid_vs_eps} visualizes these trends; the shaded bands show that NED not only achieves lower mean FID but also exhibits substantially lower cross-seed variance than KL. Additional results, including CIFAR-100 (where NED reduces FID from $98.0$ to $75.3$ at $\varepsilon{=}0.30$) and a quadrature resolution ablation for NED, are reported in Appendix~\ref{app:experiments}.

In terms of computational overhead, HD requires a single \texttt{exp} per sample and adds negligible cost. NED requires evaluating a 64-node Gauss--Legendre quadrature per sample, but this computation is fully vectorized over the batch and involves only elementary operations (exp, multiply, sum). On our setup, the per-step wall-clock overhead of NED relative to the U-Net forward--backward pass is below $1\%$; total training time for all three divergences is within ${\sim}3$h per 100K-step run on a single GPU, with no measurable difference between KL, HD, and NED.

\section{Discussion and Limitations}

This work provides a divergence-based foundation for diffusion training: the lognormal characterization of the per-step likelihood ratio yields a unified view of diffusion training as divergence-induced weighted denoising, and the experiments support the conclusion that bounded-influence divergences outperform both KL and standard robust losses under contamination.

The connection to robust statistics runs deeper than an analogy. In classical M-estimation, robustness is governed by the residual score $\psi(r)=\rho'(r)$; boundedness of $\psi$ is the standard criterion \citep{huber1964robust,hampel1986robust}. In our setting, with residual $r=\epsilon-\epsilon_\theta$ and mismatch $m=w\|r\|^2$, the residual-space score is
\[
\psi_G(r) = 2w\,h_G'(w\|r\|^2)\,r.
\]
For KL, $h_{\mathrm{KL}}'(m)=1$, so $\psi_{\mathrm{KL}}(r)=2wr$ is unbounded in $\|r\|$. For HD and NED, $h_G'(m)$ decays exponentially, making the residual-space score bounded and redescending, thereby suppressing high-mismatch samples. This parallels the role of redescending $\psi$-functions in robust regression. A full estimator-level influence-function analysis, quantifying the breakdown point of the induced training procedure, remains an open direction.

From a practitioner's perspective, the recipe is simple: replace the MSE loss with $h_G(w_t\|\epsilon{-}\epsilon_\theta\|^2)$ for a bounded-influence $G$. HD is the default when a closed-form weight is preferred; NED provides stronger suppression at the cost of a quadrature evaluation (negligible overhead). Extension to flow matching and limitations are discussed in Appendix~\ref{app:limitations}.

\bibliographystyle{plainnat}
\bibliography{references}

\newpage
\appendix
\section*{Appendix}

\section{Proof of Lemma~\ref{lem:lognormal}}
\label{app:proof_lognormal}

Conditional on $(x_0,x_t)$, the means $\tilde\mu_t=\tilde\mu_t(x_0,x_t)$ and $\mu_{\theta,t}=\mu_{\theta,t}(x_t)$ are fixed. Under matched covariance $\Sigma_t$, the log likelihood ratio between $q_t=\mathcal N(\tilde\mu_t,\Sigma_t)$ and $p_{\theta,t}=\mathcal N(\mu_{\theta,t},\Sigma_t)$ is
\[
\log r_t
= \log\frac{q_t(x_{t-1})}{p_{\theta,t}(x_{t-1})}
= -\tfrac12\|\Sigma_t^{-1/2}(x_{t-1}-\tilde\mu_t)\|^2 + \tfrac12\|\Sigma_t^{-1/2}(x_{t-1}-\mu_{\theta,t})\|^2.
\]
Write $\Delta:=\Sigma_t^{-1/2}(\tilde\mu_t-\mu_{\theta,t})$ so that $m_t=\tfrac12\|\Delta\|^2$. Under the local model kernel $p_{\theta,t}(\cdot\mid x_t)$, set $z:=\Sigma_t^{-1/2}(x_{t-1}-\mu_{\theta,t})\sim N(0,I)$. Expanding and simplifying gives $\log r_t = \Delta^\top z - \tfrac12\|\Delta\|^2$. Since $\Delta^\top z\sim N(0,\|\Delta\|^2)=N(0,2m_t)$, we obtain $\log r_t \sim N(-m_t, 2m_t)$ conditional on $(x_0,x_t)$.

The moments follow from standard lognormal calculations: $\E[r_t\mid x_0,x_t]=e^{-m_t+m_t}=1$ and $\mathrm{Var}(r_t\mid x_0,x_t)=e^{2m_t}-1$. \qed

\section{Proof of Theorem~\ref{thm:marginal_to_path}}
\label{app:proof_marginal_to_path}

The map \(x_{0:T} \mapsto x_0\) is a measurable projection from path space to observation space. The marginals of \(Q\) and \(P_\theta\) under this projection are exactly \(g\) and \(f_\theta\), respectively. Since \(f\)-divergences are monotone under measurable maps, we obtain
\[
D_G(g\|f_\theta)
\le
D_G(Q\|P_\theta).
\]

For equality, note that equality in the data-processing inequality holds when the likelihood ratio \(dQ/dP_\theta\) depends only on the projected variable \(x_0\). A sufficient condition is that the conditional latent laws agree given \(x_0\):
\[
Q(x_{1:T}\mid x_0)=P_\theta(x_{1:T}\mid x_0)
\quad
\text{for } g\text{-a.e. }x_0.
\]
Under this condition, the path likelihood ratio reduces to the marginal likelihood ratio, so
\[
D_G(g\|f_\theta)=D_G(Q\|P_\theta).
\]
The final claim follows: if $\theta_0$ gives an exact lifted model ($P_{\theta_0}=Q$), the conditional-path equality holds and the bound is tight. \qed

\section{Proof of Theorem~\ref{thm:path_to_local_clean}}
\label{app:proof_path_to_local}

Since \(G\in C^3\) near \(0\) with \(G(0)=G'(0)=0\) and \(G''(0)=1\), there exist constants \(\eta>0\), \(C>0\)
such that for all \(|x|\le \eta\),
\[
G(x)=\tfrac{1}{2}x^2+\mathcal R(x),
\qquad
|\mathcal R(x)|\le C|x|^3.
\]
Because \(a_T\to 0\), assumptions (A1), (A3), (A4), and (A5) imply
\[
S_T \to 0 \quad \text{in }L^2,\qquad
B_T \to 0 \quad \text{in }L^2,
\]
hence
\[
S_T+B_T \to 0 \quad \text{in probability}.
\]
The tail integrability assumption ensures that the contribution from the event \(\{|S_T+B_T|>\eta\}\) is \(o(a_T)\). On the complementary event, the Taylor expansion applies. By (A3) and (A5),
\[
\mathbb{E}|S_T+B_T|^3
\le
C\bigl(\mathbb{E}|S_T|^3+\mathbb{E}|B_T|^3\bigr)
=
o(a_T).
\]
The quadratic term on the tail is also controlled: $\mathbb E[(S_T+B_T)^2\,\mathbf 1\{|S_T+B_T|>\eta\}] \le \eta^{-1}\mathbb E|S_T+B_T|^3 = o(a_T)$. Hence
\[
\mathbb{E}[\Phi_T]
=
\tfrac{1}{2}\mathbb{E}[(S_T+B_T)^2]
+
o(a_T).
\]
Expanding the square gives
\[
\mathbb{E}[(S_T+B_T)^2]
=
\mathbb{E}[S_T^2] + 2\mathbb{E}[S_TB_T] + \mathbb{E}[B_T^2]
=
\mathbb{E}[S_T^2] + o(a_T)
\]
by (A4). Hence
\[
\mathbb{E}[\Phi_T]
=
\tfrac{1}{2}\mathbb{E}[S_T^2] + o(a_T).
\]

On the other hand, for each \(t\),
\[
G(\delta_t)=\tfrac{1}{2}\delta_t^2+\mathcal R(\delta_t),
\qquad
|\mathcal R(\delta_t)|\le C|\delta_t|^3
\]
for \(|\delta_t|\) sufficiently small. Similarly, $\sum_t \mathbb E[\delta_t^2\,\mathbf 1\{|\delta_t|>\eta\}] \le \eta^{-1}\sum_t\mathbb E|\delta_t|^3 = o(a_T)$ by~(A6), and the per-step tail integrability assumption controls the $G$ contribution on the tails. Summing yields
\[
\mathbb{E}[\Psi_T]
=
\tfrac{1}{2}\sum_{t=1}^T \mathbb{E}[\delta_t^2]
+
o(a_T)
\]
by (A6). Finally, assumption (A2) gives
\[
\mathbb{E}[S_T^2]
=
\sum_{t=1}^T \mathbb{E}[\delta_t^2] + o(a_T),
\]
so
\[
\mathbb{E}[\Phi_T]
=
\mathbb{E}[\Psi_T] + o(a_T).
\]
\qed

\paragraph{Interpretation of conditions (A1)--(A6).}
Conditions (A1)--(A6) are abstract moment conditions on the increments $\delta_t$ and their partial sums. In the context of diffusion models, they admit a natural interpretation as Lindeberg-type regularity conditions ensuring that the product likelihood ratio $R_T = \prod_t(1+\delta_t)$ is well approximated by its linearization $1 + S_T$.

Conditions (A1) and (A2) together require that $\E[S_T^2]$ is $O(a_T)$ and that the variance of $S_T$ is dominated by the sum of individual variances $\sum_t \E[\delta_t^2]$, up to $o(a_T)$ error. In the diffusion setting, (A2) holds exactly when the increments are pairwise uncorrelated, which is precisely what the orthogonality assumption (D2) provides.

Condition (A3) controls the third moment of the sum $S_T$, ensuring that $S_T$ does not have heavy tails. Together with (A1), this is a standard Lindeberg-type condition: it guarantees that no single increment dominates the sum, which is supported by the moment conditions in~(D1).

Conditions (A4) and (A5) bound the cross-interaction and higher moments of the product remainder $B_T = R_T - 1 - S_T$. Since $B_T$ captures the nonlinear terms in the product $\prod_t(1+\delta_t)$, these conditions ensure that the product-to-sum approximation $R_T \approx 1 + S_T$ is valid in $L^2$ and $L^3$. When the per-step mismatches are uniformly small (D1), the remainder $B_T$ is of order $O(\sum_{s<t} \delta_s \delta_t)$, and its moments are controlled by those of the individual increments.

Finally, (A6) requires $\sum_t \E|\delta_t|^3 = o(a_T)$, which ensures that the per-step Taylor remainders $\mathcal R(\delta_t)$ are negligible in aggregate. Under (D1), the lognormal structure of Lemma~\ref{lem:lognormal} gives $\E[|\delta_t|^3\mid x_0,x_t] = O(m_t^{3/2})$, so $\sum_t \E|\delta_t|^3 = O(\sum_t \E[m_t^{3/2}]) = o(a_T)$ by the moment condition in~(D1).

In summary, (A1)--(A3) and (A6) are Lindeberg-type conditions on the sum $S_T$, while (A4)--(A5) control the product remainder $B_T$. In the diffusion specialization, some of these conditions follow from the lognormal moment bounds (specifically (A1), (A2), and (A6) from (D1) and (D2)), while the third-moment condition (A3) and the product-remainder conditions (A4)--(A5) are imposed explicitly via (D3) and (D4) in Corollary~\ref{cor:diffusion_clean}.

\section{Proof of Corollary~\ref{cor:diffusion_clean}}
\label{app:proof_corollary}

All expectations below are under $\E_{\mathrm{loc}}$ (the auxiliary local-product construction defined in Corollary~\ref{cor:diffusion_clean}). By Lemma~\ref{lem:lognormal}, conditional on $(x_0,x_t)$, $\log r_t \sim N(-m_t,2m_t)$ under the local model kernel $p_{\theta,t}(\cdot\mid x_t)$. Defining $h_G(m):=\E[G(U_m-1)]$ with $\log U_m\sim N(-m,2m)$, the tower property gives $\E_{\mathrm{loc}}[G(\delta_t)] = \E[h_G(m_t)]$, and hence
\[
\E_{\mathrm{loc}}[\Psi_T]=\sum_{t=1}^T \E[h_G(m_t)].
\]
Standard lognormal moment calculations give
\[
\E_{\mathrm{loc}}[\delta_t^2\mid x_0,x_t]=e^{2m_t}-1 = 2m_t+O(m_t^2).
\]
By the uniform-smallness condition $\max_t m_t \le c_T\to 0$ in~(D1), the local expansion $\E_{\mathrm{loc}}[|\delta_t|^3\mid x_0,x_t] = O(m_t^{3/2})$ holds uniformly over all steps (the $O(\cdot)$ constant depends only on $c_T$, which is bounded).
Taking expectations gives
\[
\sum_{t=1}^T \E_{\mathrm{loc}}[\delta_t^2]
=
2a_T + O\!\left(\sum_{t=1}^T \E[m_t^2]\right)
=
2a_T + o(a_T)
=
O(a_T),
\]
by the moment condition $\sum_t \E[m_t^2]=o(a_T)$ in~(D1), and
\[
\sum_{t=1}^T \E_{\mathrm{loc}}|\delta_t|^3
=
O\!\left(\sum_{t=1}^T \E[m_t^{3/2}]\right)
=
o(a_T),
\]
by the moment condition $\sum_t \E[m_t^{3/2}]=o(a_T)$ in~(D1).

Next, assumption (D2) directly gives $\E_{\mathrm{loc}}[\delta_s\delta_t]=0$ for $s\ne t$. Hence
\[
\E_{\mathrm{loc}}[S_T^2]
=
\sum_{t=1}^T \E_{\mathrm{loc}}[\delta_t^2].
\]
The argument above verifies (A1), (A2), and (A6) of Theorem~\ref{thm:path_to_local_clean} from (D1) and (D2). The third-moment condition (A3), $\E_{\mathrm{loc}}|S_T|^3=o(a_T)$, is imposed directly as assumption~(D3). The tail integrability condition is assumed explicitly in Corollary~\ref{cor:diffusion_clean}; per-step tails are natural for KL, HD, and NED because their generators have at most linear-times-logarithmic growth in $U_m$ (see Appendix~\ref{app:proof_local}), and the super-exponential decay of $P(|U_m-1|>\eta)$ as $m\to 0$ ensures the tail conditions hold. The product-remainder conditions (A4)--(A5) are imposed as~(D4).
Applying Theorem~\ref{thm:path_to_local_clean} gives
\[
\E_{\mathrm{loc}}[G(R_T-1)]
=
\sum_{t=1}^T \E_{\mathrm{loc}}[G(\delta_t)] + o(a_T).
\]
Since $\E_{\mathrm{loc}}[G(\delta_t)]=\E[h_G(m_t)]$,
the conclusion $\E_{\mathrm{loc}}[G(R_T{-}1)] = \sum_t \E[h_G(m_t)] + o(a_T)$ follows. \qed

\section{Proof of Theorem~\ref{thm:local_convergence}}
\label{app:proof_local}

Let \(\delta = U_m - 1\). Since $G\in C^3$ near $0$ with $G(0)=G'(0)=0$, $G''(0)=1$, Taylor expansion gives
\[
G(\delta)
=
\frac{1}{2}\delta^2 + O(|\delta|^3).
\]
The stated tail condition $\E[|G(U_m-1)|\,\mathbf{1}\{|U_m-1|>\eta\}]=O(m^{3/2})$ directly controls the contribution outside the Taylor neighborhood. Note that $U_m$ is lognormal, so all its polynomial moments are finite; however, its moment generating function diverges ($\E[e^{cU_m}]=\infty$ for $c>0$), so generators with exponential or faster growth require case-by-case verification of the tail condition. Taking expectation on the Taylor region,
\[
h_G(m)
=
\frac{1}{2}\mathbb E[\delta^2]
+
O\!\left(\mathbb E|\delta|^3\right).
\]
By Lemma~\ref{lem:lognormal}, \(\log U_m \sim N(-m,2m)\), so standard lognormal moment calculations give
\[
\mathbb E[\delta^2]=e^{2m}-1=2m+O(m^2),
\qquad
\mathbb E|\delta|^3=O(m^{3/2}).
\]
Substituting yields
\[
h_G(m)=m+O(m^{3/2}).
\]

We now verify the sharper $O(m^2)$ rate for KL, HD, and NED. Each generator is $C^\infty$ near $0$, so we can write $G(\delta)=\frac{1}{2}\delta^2+\frac{1}{6}G^{(3)}(0)\delta^3+\frac{1}{24}G^{(4)}(0)\delta^4+R_5(\delta)$ where $|R_5(\delta)|\le C|\delta|^5$ for $|\delta|\le\eta$. Using the exact lognormal moments $\E[\delta^2]=2m+2m^2+O(m^3)$, $\E[\delta^3]=12m^2+O(m^3)$, $\E[\delta^4]=12m^2+O(m^3)$, and $\E|\delta|^5=O(m^{5/2})$, we obtain $h_G(m)=m+O(m^2)$ provided the tail condition $\E[|G(U_m-1)|\,\mathbf{1}\{|U_m-1|>\eta\}]=O(m^2)$ holds. We verify this via a lognormal Chernoff bound: since $U_m\to 1$ sharply as $m\to 0$, for fixed $\eta>0$, $P(|U_m-1|>\eta)$ decays like $\exp(-c_\eta/m)$. For the three generators: KL has $|G(U_m{-}1)|\le C\,U_m\log U_m$; HD has $|G(U_m{-}1)|\le C\,U_m$; NED has $|G(U_m{-}1)|\le C\,U_m$. In each case, the growth is at most linear-times-logarithmic in $U_m$, and the super-exponential decay of the tail probability dominates, so the $O(m^2)$ tail condition is satisfied.
\qed

\section{Proof of Theorem~\ref{thm:global_monotonicity}}
\label{app:proof_monotonicity}

Define
\[
\phi(u):=G(u-1),\qquad u>0.
\]
Since \(G\) is convex on \([-1,\infty)\), \(\phi\) is convex on \((0,\infty)\). Also,
\[
\phi(1)=G(0)=0,\qquad \phi'(1)=G'(0)=0.
\]

First, by convexity of \(G\),
\[
G(\delta)\ge G(0)+G'(0)\delta =0
\qquad \text{for all } \delta\ge -1.
\]
Hence
\[
h_G(m)=\mathbb{E}[G(U_m-1)]\ge 0.
\]
When \(m=0\), we have \(U_0=1\) almost surely, so
\[
h_G(0)=G(0)=0.
\]
Now let \((B_t)_{t\ge 0}\) be a standard Brownian motion and define
\[
M_t:=\exp(\sqrt{2}\,B_t-t),\qquad t\ge 0.
\]
Then \(M_t\) is a positive martingale, and for each fixed \(m\ge 0\),
\[
M_m \overset{d}= \exp(\sqrt{2m}\,Z-m)=U_m.
\]
Therefore,
\[
h_G(m)=\mathbb{E}[\phi(M_m)].
\]

Let \(0\le s\le t\). Since \(M_t\) is a martingale and \(\phi\) is convex, conditional Jensen's inequality gives
\[
\mathbb{E}[\phi(M_t)\mid \mathcal{F}_s]
\ge
\phi\!\bigl(\mathbb{E}[M_t\mid \mathcal{F}_s]\bigr)
=
\phi(M_s).
\]
Thus \((\phi(M_t))_{t\ge 0}\) is a submartingale. Taking expectations yields
\[
\mathbb{E}[\phi(M_t)]\ge \mathbb{E}[\phi(M_s)].
\]
Equivalently,
\[
h_G(t)\ge h_G(s),
\]
so \(h_G\) is nondecreasing on \([0,\infty)\).

Finally, assume \(G\) is strictly convex. Then \(\phi\) is strictly convex. For \(m>0\), the random variable \(U_m\) is nondegenerate and satisfies \(\mathbb{E}[U_m]=1\). Hence, by strict Jensen,
\[
h_G(m)=\mathbb{E}[\phi(U_m)]
>
\phi(\mathbb{E}[U_m])
=
\phi(1)
=
0.
\]
Therefore \(h_G(m)>0\) for all \(m>0\). \qed

\section{Proof of Proposition~\ref{prop:ned_decay}}
\label{app:ned_decay}

Since $\E[U_m]=1$, we have $h_{\mathrm{NED}}(m)=e\,\E[e^{-U_m}]-1$. Differentiating under the integral sign via $\partial U_m/\partial m = U_m(-1+Z/\sqrt{2m})$ gives
\[
h'_{\mathrm{NED}}(m)
= e\,\E\!\left[e^{-U_m}\,U_m\!\left(1-\tfrac{Z}{\sqrt{2m}}\right)\right].
\]
\textbf{Girsanov shift.} Since $U_m=e^{-m+\sqrt{2m}Z}$ with $Z\sim N(0,1)$, the change of measure $\E[U_m\,f(Z)]=\E[f(Z+\sqrt{2m})]$ transforms the expectation to
\[
h'_{\mathrm{NED}}(m)
= \frac{e}{\sqrt{2m}}\,\E\!\left[|Z|\,e^{-\exp(m-\sqrt{2m}|Z|)}\,\mathbf{1}\{Z<0\}\right] + o(m^{-1/2}e^{-m/4}),
\]
where the $Z\ge 0$ contribution is super-exponentially small and hence negligible at the $m^{-1/2}e^{-m/4}$ scale.

\textbf{Laplace analysis.} Substitute $|Z|=\sqrt{m/2}+s$ (the transition point where $\exp(m-\sqrt{2m}|Z|)=1$):
\[
h'_{\mathrm{NED}}(m)
\sim \frac{e}{\sqrt{2m}}\cdot\sqrt{\tfrac{m}{2}}\,\phi\!\left(\sqrt{\tfrac{m}{2}}\right)
\int_{-\infty}^{\infty}
e^{-e^{-\sqrt{2m}\,s}}\,e^{-s\sqrt{m/2}}\,ds.
\]
Set $u=e^{-\sqrt{2m}\,s}$, giving $e^{-s\sqrt{m/2}}=u^{1/2}$ and $ds=-du/(u\sqrt{2m})$. The integral becomes $\frac{1}{\sqrt{2m}}\int_0^\infty e^{-u}u^{-1/2}\,du = \frac{\Gamma(1/2)}{\sqrt{2m}}=\frac{\sqrt{\pi}}{\sqrt{2m}}$.
Combining all factors and using $\phi(\sqrt{m/2})=\frac{1}{\sqrt{2\pi}}e^{-m/4}$:
\[
h'_{\mathrm{NED}}(m)
\sim
\frac{e}{\sqrt{2m}}\cdot\sqrt{\tfrac{m}{2}}\cdot\frac{e^{-m/4}}{\sqrt{2\pi}}\cdot\frac{\sqrt{\pi}}{\sqrt{2m}}
= \frac{e}{4\sqrt{m}}\,e^{-m/4}.
\]
\qed

\section{Extended Experimental Results}
\label{app:experiments}

All experiments use the same 35M-parameter DDPM U-Net architecture described in Section~\ref{sec:experiments}, trained for 100K steps on a single NVIDIA RTX 5090 GPU (${\sim}3$h per run). FID is computed with 5K generated samples against the clean test split via a 50-step DDIM sampler. Training uses Adam at $\mathrm{lr}=2{\times}10^{-4}$, batch size 128, bfloat16 mixed-precision, and gradient clipping at norm~1.0.

\paragraph{CIFAR-10 multi-seed protocol.}
Three divergences (KL, HD, NED) $\times$ four contamination levels ($\varepsilon\in\{0.00, 0.10, 0.20, 0.30\}$) $\times$ three seeds (42, 123, 456) $=$ 36 runs (${\sim}108$ GPU-hours). Results are reported in Table~\ref{tab:cifar_results} (main text).

\paragraph{CIFAR-100 results.}
Table~\ref{tab:cifar100} reports single-seed CIFAR-100 FID scores. The pattern is consistent with CIFAR-10: NED achieves the best FID under heavy contamination ($75.3$ vs.\ $98.0$ for KL), while all divergences are comparable on clean data.

\begin{table}[h]
\centering
\caption{CIFAR-100 FID (single seed, 100K steps).}
\label{tab:cifar100}
\begin{tabular}{lccc}
\toprule
$\varepsilon$ & KL & HD & NED \\
\midrule
$0.00$ & $\bm{35.0}$ & $35.8$ & $35.6$ \\
$0.10$ & $46.4$ & $48.0$ & $\bm{44.7}$ \\
$0.20$ & $64.3$ & $\bm{53.9}$ & $56.9$ \\
$0.30$ & $98.0$ & $99.4$ & $\bm{75.3}$ \\
\bottomrule
\end{tabular}
\end{table}

\paragraph{NED quadrature ablation.}
Since $h_{\mathrm{NED}}'(m)$ has no closed form, it is approximated by Gauss--Legendre quadrature over the lognormal variable $U_m$ (Lemma~\ref{lem:lognormal}). Table~\ref{tab:ablations} varies the number of quadrature nodes on clean CIFAR-10.

\begin{table}[h]
\centering
\caption{NED quadrature resolution on clean CIFAR-10 (single seed, 100K steps).}
\label{tab:ablations}
\begin{tabular}{lc}
\toprule
Quadrature nodes & FID \\
\midrule
16 & $\bm{25.7}$ \\
32 & $30.0$ \\
64 & $29.7$ \\
128 & $29.7$ \\
\bottomrule
\end{tabular}
\end{table}

Performance is broadly stable across quadrature resolutions; the variation (FID 25.7--30.0) is within the cross-seed noise observed in Table~\ref{tab:cifar_results}. All experiments in this paper use 64 nodes as a conservative default to ensure numerical stability across all mismatch levels.

\paragraph{Additional figures.}

\begin{figure}[h]
\centering
\includegraphics[width=0.85\textwidth]{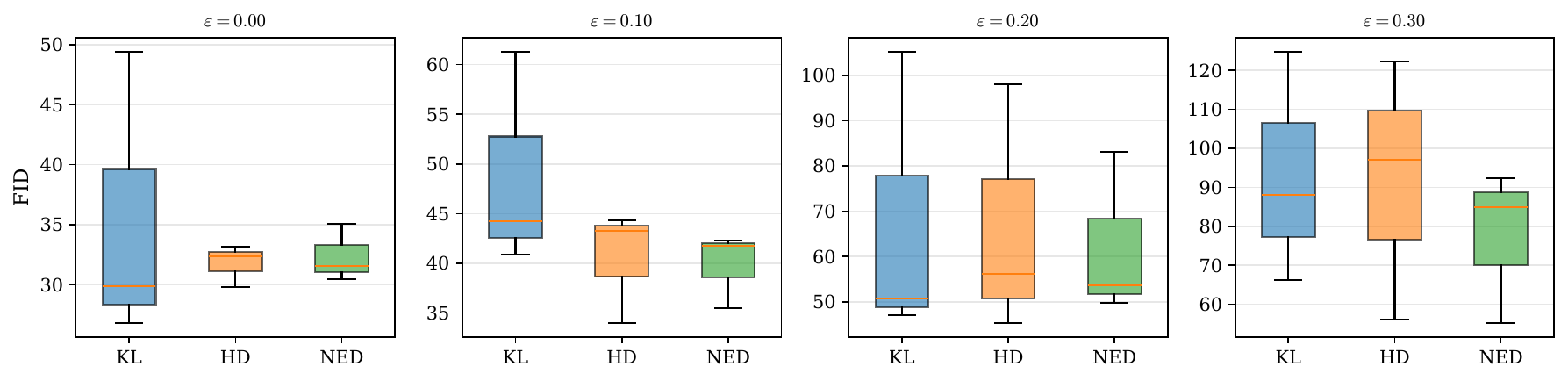}
\caption{Box plots of CIFAR-10 FID across 3 seeds at each contamination level. HD and NED exhibit tighter interquartile ranges than KL, especially on clean data, indicating that bounded-influence objectives stabilize training.}
\label{fig:multiseed_box}
\end{figure}

\begin{figure}[h]
\centering
\includegraphics[width=\textwidth]{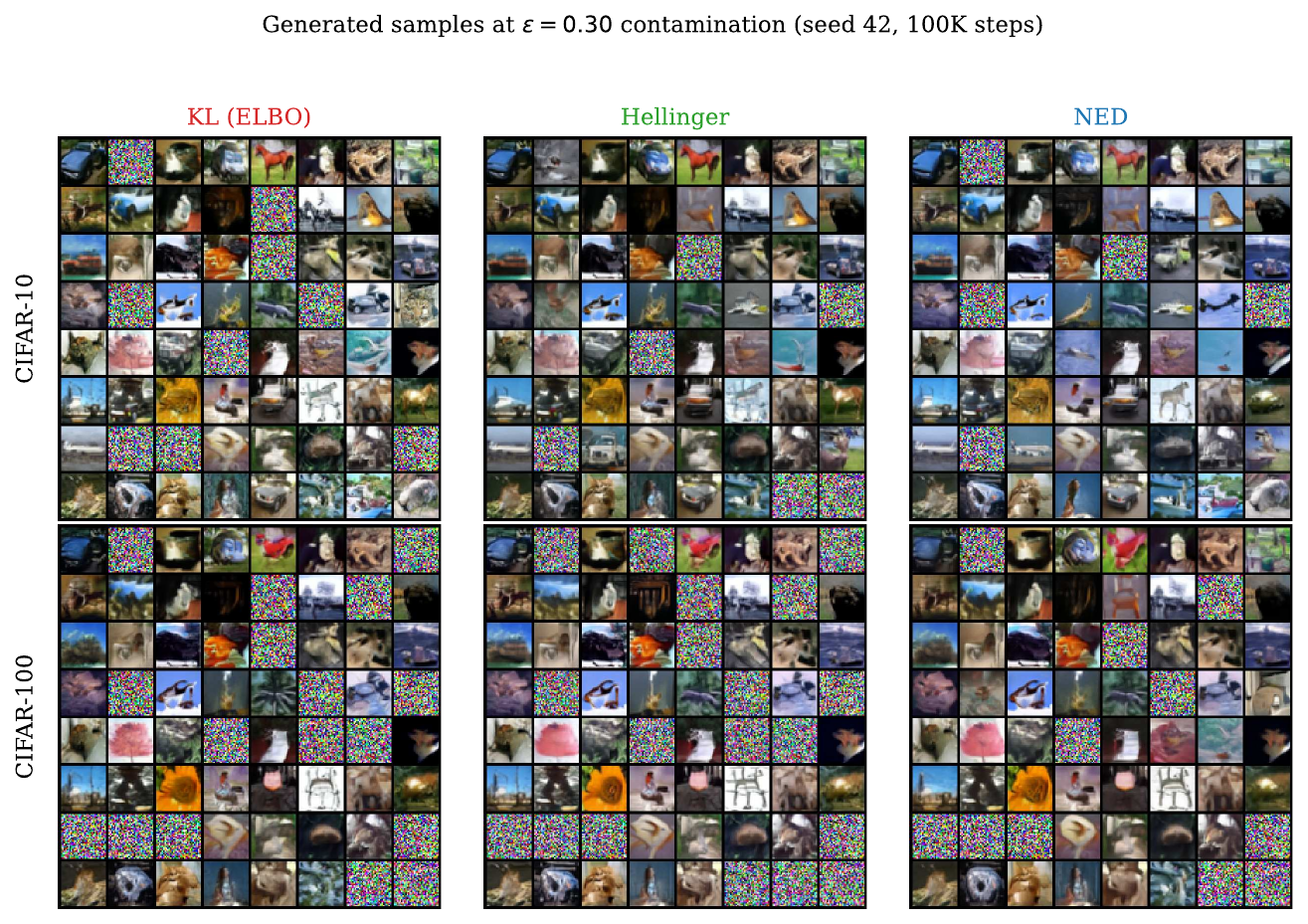}
\caption{Generated samples at $\varepsilon{=}0.30$ contamination (seed 42, 100K steps). \textbf{Top:} CIFAR-10. \textbf{Bottom:} CIFAR-100. KL samples exhibit visible contamination artifacts (mosaic noise patches), while HD and NED produce cleaner images, consistent with their bounded-influence weighting.}
\label{fig:samples_grid}
\end{figure}

\begin{figure}[h]
\centering
\includegraphics[width=0.55\textwidth]{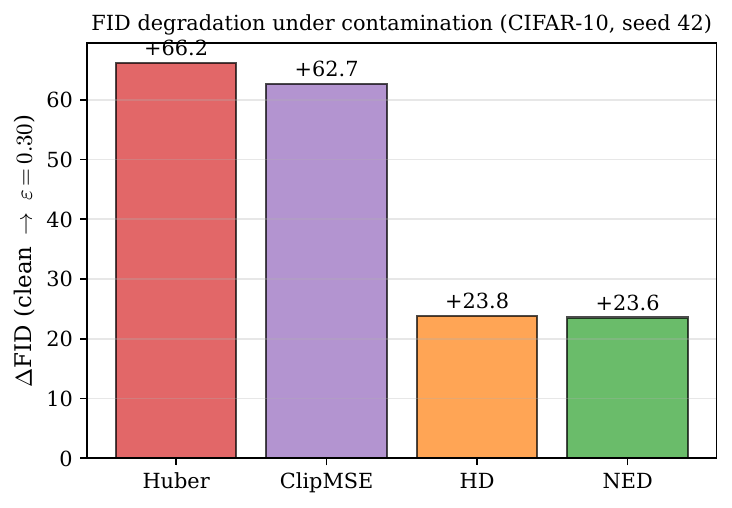}
\caption{FID degradation ($\Delta$FID from clean to $\varepsilon{=}0.30$) for standard robust losses vs.\ divergence-induced objectives on CIFAR-10 (single seed). NED degrades least ($+23.5$), while Huber and clipped MSE degrade substantially.}
\label{fig:baseline_bars}
\end{figure}

\section{Limitations and Broader Impact}
\label{app:limitations}

\paragraph{Limitations.}
The local conditional objective is a training surrogate motivated by the path-space upper bound, but is not itself a proven upper bound on the marginal divergence $D_G(g\|f_\theta)$ for general (non-KL) f-divergences; the gap between the surrogate and the marginal divergence remains an open question. The asymptotic product-to-sum reduction (Theorem~\ref{thm:path_to_local_clean}) is first-order in the local mismatch; finite-mismatch corrections remain an open direction. The robustness claims are empirically supported by the contamination experiments but are not accompanied by formal contamination-risk bounds or breakdown-point guarantees. While our multi-seed CIFAR-10 and single-seed CIFAR-100 experiments demonstrate the robustness advantage of bounded-influence divergences, the cross-seed variance is substantial (particularly at high contamination), and evaluation on larger-scale datasets would strengthen the empirical evidence. The framework assumes Gaussian reverse kernels (DDPM variational posteriors) with matched covariance; extending to non-Gaussian or learnable-covariance settings would broaden applicability. Under the Gaussian reverse-kernel structure, the framework extends naturally to flow matching by applying $h_G$ to the velocity-field residual $\|v_\theta-v^*\|^2$; empirical validation of this extension is left to future work.

\paragraph{Broader impact.}
Diffusion models are widely used for image and audio generation, and improvements to their training objectives may amplify both beneficial applications (e.g., scientific visualization, creative tools) and potential misuse (e.g., deepfakes, disinformation). Our robust training framework may partially mitigate data-poisoning attacks by downweighting corrupted samples, but it does not address misuse of the generated content itself. We encourage practitioners to pair improved training methods with appropriate content safeguards.

\end{document}